\documentclass[10pt,twocolumn,letterpaper]{article}

\usepackage{cvpr}
\usepackage{times}
\usepackage{epsfig}
\usepackage{graphicx}
\usepackage{amsmath}
\usepackage{amssymb}
\usepackage{color}
\usepackage{subfigure}
\usepackage{multirow}

\usepackage[pagebackref=true,breaklinks=true,letterpaper=true,colorlinks,bookmarks=false]{hyperref}

\cvprfinalcopy 

\def\cvprPaperID{2265} 

\ifcvprfinal\fi
\begin{document}

\title{MobileNets: Efficient Convolutional Neural Networks for Mobile Vision Applications}

\author{Andrew G. Howard
\qquad
Menglong Zhu
\qquad
Bo Chen
\qquad
Dmitry Kalenichenko\\
Weijun Wang
\qquad
Tobias Weyand
\qquad
Marco Andreetto
\qquad
Hartwig Adam\\
\vspace{0.2em}\\
Google Inc.\\
{\tt\small \{howarda,menglong,bochen,dkalenichenko,weijunw,weyand,anm,hadam\}@google.com}
}

\maketitle

\begin{abstract}
   We present a class of efficient models called MobileNets for mobile and embedded vision applications. MobileNets are based on a streamlined architecture that uses depthwise separable convolutions to build light weight deep neural networks. We introduce two simple global hyper-parameters that efficiently trade off between latency and accuracy. These hyper-parameters allow the model builder to choose the right sized model for their application based on the constraints of the problem. We present extensive experiments on resource and accuracy tradeoffs and show strong performance compared to other popular models on ImageNet classification. We then demonstrate the effectiveness of MobileNets across a wide range of applications and use cases including object detection, finegrain classification, face attributes and large scale geo-localization.
\end{abstract}

\section{Introduction}

Convolutional neural networks have become ubiquitous in computer vision ever since AlexNet \cite{krizhevsky2012imagenet} popularized deep convolutional neural networks by winning the ImageNet Challenge: ILSVRC 2012 \cite{russakovsky2015imagenet}. The general trend has been to make deeper and more complicated networks in order to achieve higher accuracy \cite{simonyan2014very,szegedy2015rethinking,szegedy2016inception,he2015deep}. However, these advances to improve accuracy are not necessarily making networks more efficient with respect to size and speed. In many real world applications such as robotics, self-driving car and augmented reality, the recognition tasks need to be carried out in a timely fashion on a computationally limited platform.

This paper describes an efficient network architecture and a set of two hyper-parameters in order to build very small, low latency models that can be easily matched to the design requirements for mobile and embedded vision applications. Section \ref{sec:prior} reviews prior work in building small models. Section \ref{sec:mobilenet} describes the MobileNet architecture and two hyper-parameters width multiplier and resolution multiplier to define smaller and more efficient MobileNets. Section \ref{sec:exp} describes experiments on ImageNet as well a variety of different applications and use cases. Section \ref{sec:conclusion} closes with a summary and conclusion.

\begin{figure*}
\centering
  \includegraphics[width=2\columnwidth]{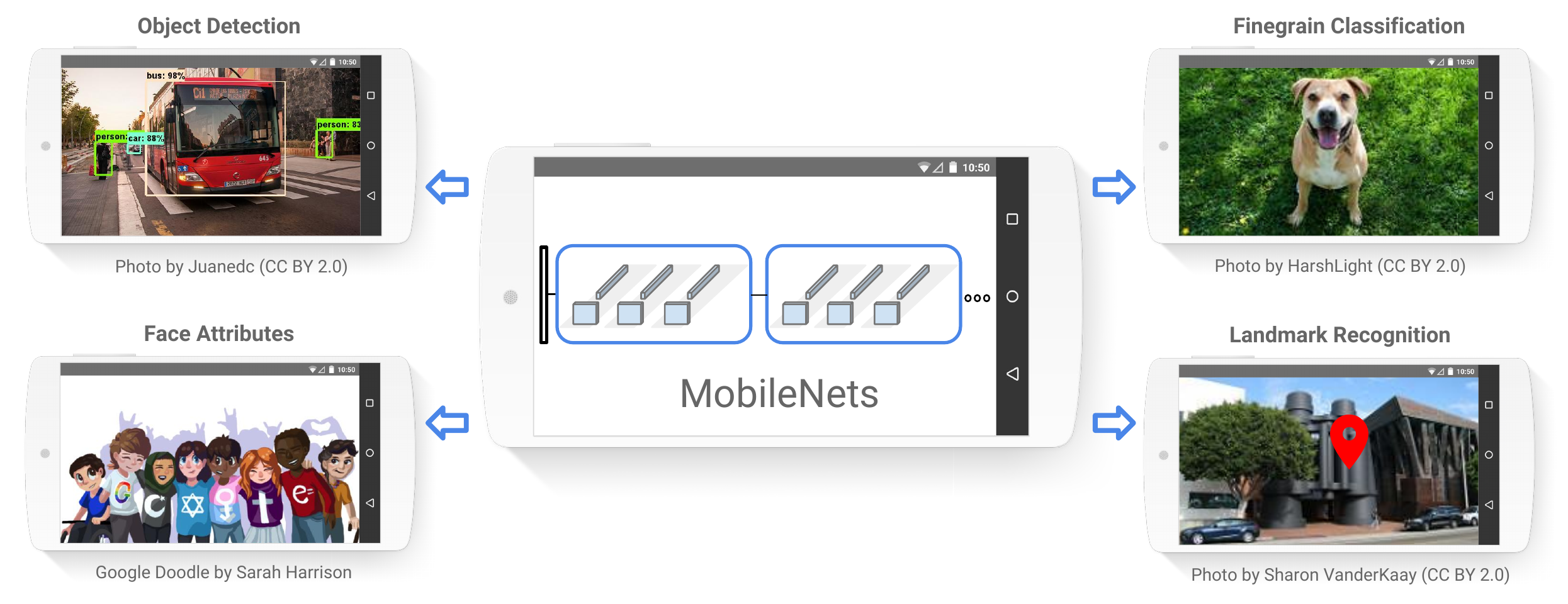}
  \caption{MobileNet models can be applied to various recognition tasks for efficient on device intelligence.}
  \label{fig:conv_layers}
\end{figure*}

\section{Prior Work} \label{sec:prior}
There has been rising interest in building small and efficient neural networks in the recent literature, e.g. \cite{jin2014flattened,wang2016factorized,iandola2016squeezenet,wu2015quantized,rastegari2016xnor}. Many different approaches can be generally categorized into either compressing pretrained networks or training small networks directly. This paper proposes a class of network architectures that allows a model developer to specifically choose a small network that matches the resource restrictions (latency, size) for their application. MobileNets primarily focus on optimizing for latency but also yield small networks. Many papers on small networks focus only on size but do not consider speed.

MobileNets are built primarily from depthwise separable convolutions initially introduced in \cite{sifre2014rigid} and subsequently used in Inception models \cite{ioffe2015batch} to reduce the computation in the first few layers. Flattened networks \cite{jin2014flattened} build a network out of fully factorized convolutions and showed the potential of extremely factorized networks. Independent of this current paper, Factorized Networks\cite{wang2016factorized} introduces a similar factorized convolution as well as the use of topological connections. Subsequently, the Xception network \cite{chollet2016deep} demonstrated how to scale up depthwise separable filters to out perform Inception V3 networks. Another small network is Squeezenet \cite{iandola2016squeezenet} which uses a bottleneck approach to design a very small network. Other reduced computation networks include structured transform networks \cite{sindhwani2015structured} and deep fried convnets \cite{yang2015deep}.

A different approach for obtaining small networks is shrinking, factorizing or compressing pretrained networks. Compression based on product quantization \cite{wu2015quantized}, hashing \cite{chen2015compressing}, and pruning, vector quantization and Huffman coding \cite{han2015deep} have been proposed in the literature. Additionally various factorizations have been proposed to speed up pretrained networks \cite{jaderberg2014speeding, lebedev2014speeding}. Another method for training small networks is distillation \cite{hinton2015distilling} which uses a larger network to teach a smaller network. It is complementary to our approach and is covered in some of our use cases in section \ref{sec:exp}. Another emerging approach is low bit networks \cite{courbariaux2014training, rastegari2016xnor, hubara2016quantized}.

\section{MobileNet Architecture} \label{sec:mobilenet}

In this section we first describe the core layers that MobileNet is built on which are depthwise separable filters. We then describe the MobileNet network structure and conclude with descriptions of the two model shrinking hyper-parameters width multiplier and resolution multiplier.

\subsection{Depthwise Separable Convolution}

\begin{figure}[t]
\centering
\subfigure[Standard Convolution Filters]{
  \includegraphics[width=.45\textwidth]{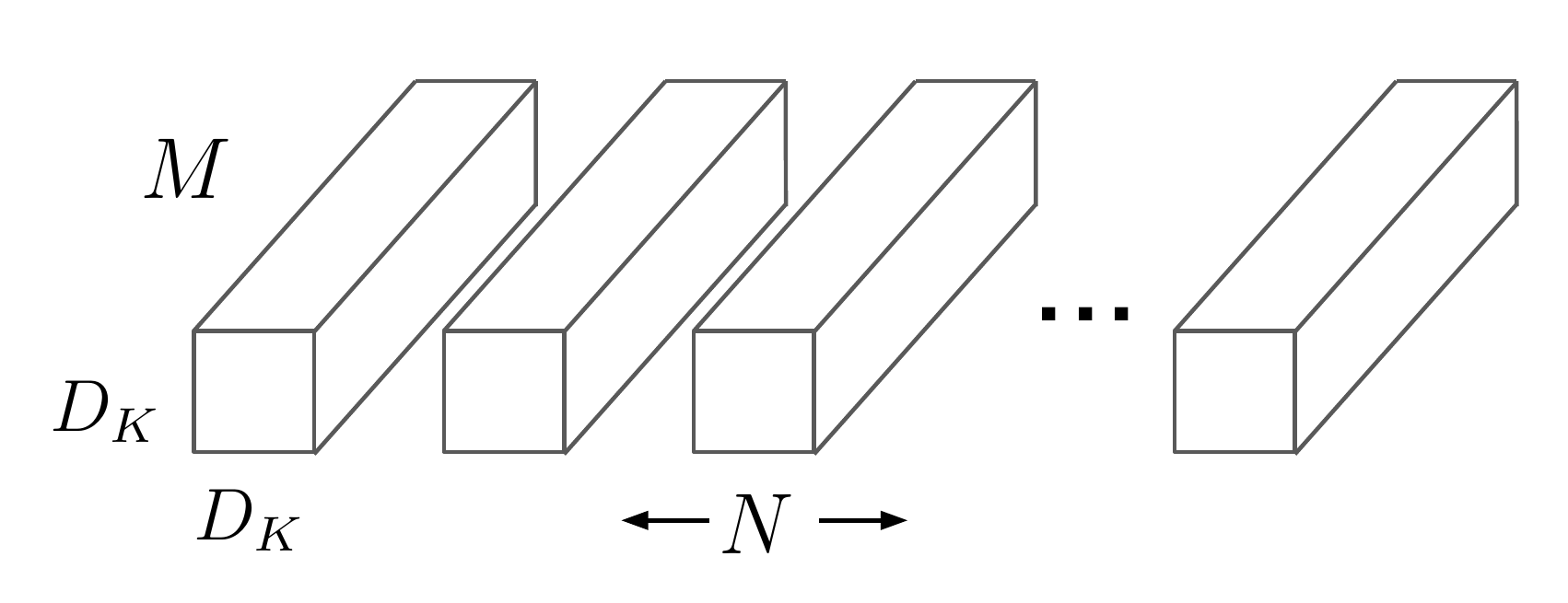}
  \label{fig:dw_conv_a}
}
\subfigure[Depthwise Convolutional Filters]{
  \includegraphics[width=.45\textwidth]{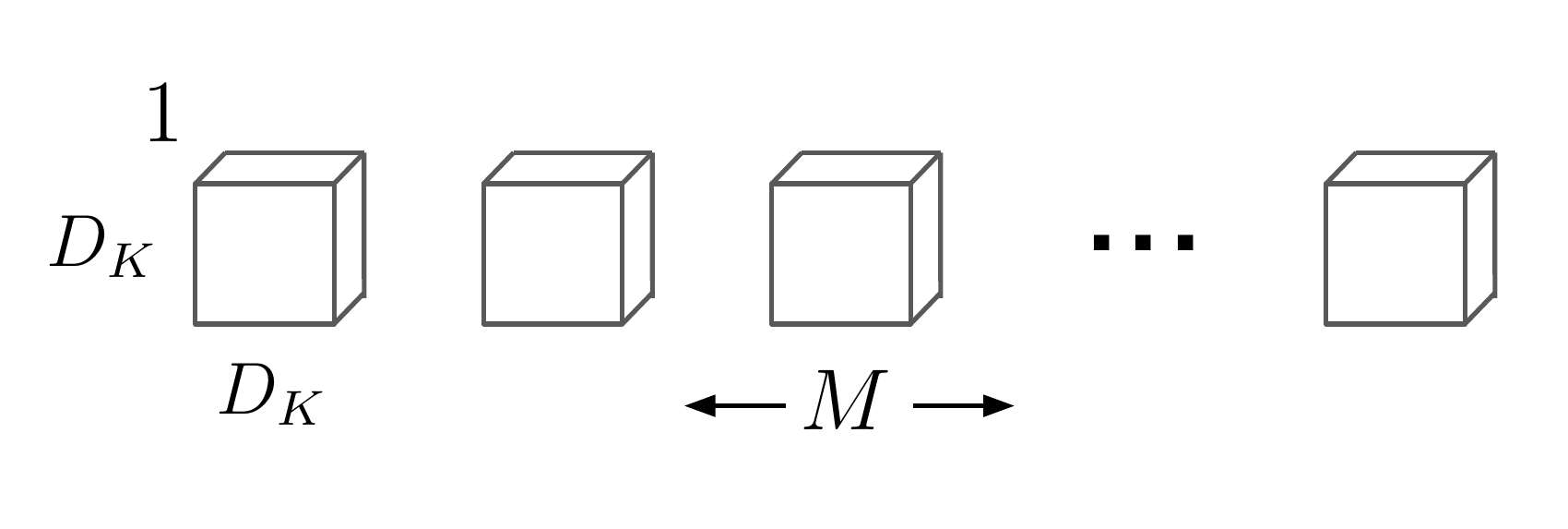}
  \label{fig:dw_conv_b}
}
\subfigure[$1 \times 1$ Convolutional Filters called Pointwise Convolution in the context of Depthwise Separable Convolution]{
  \includegraphics[width=.45\textwidth]{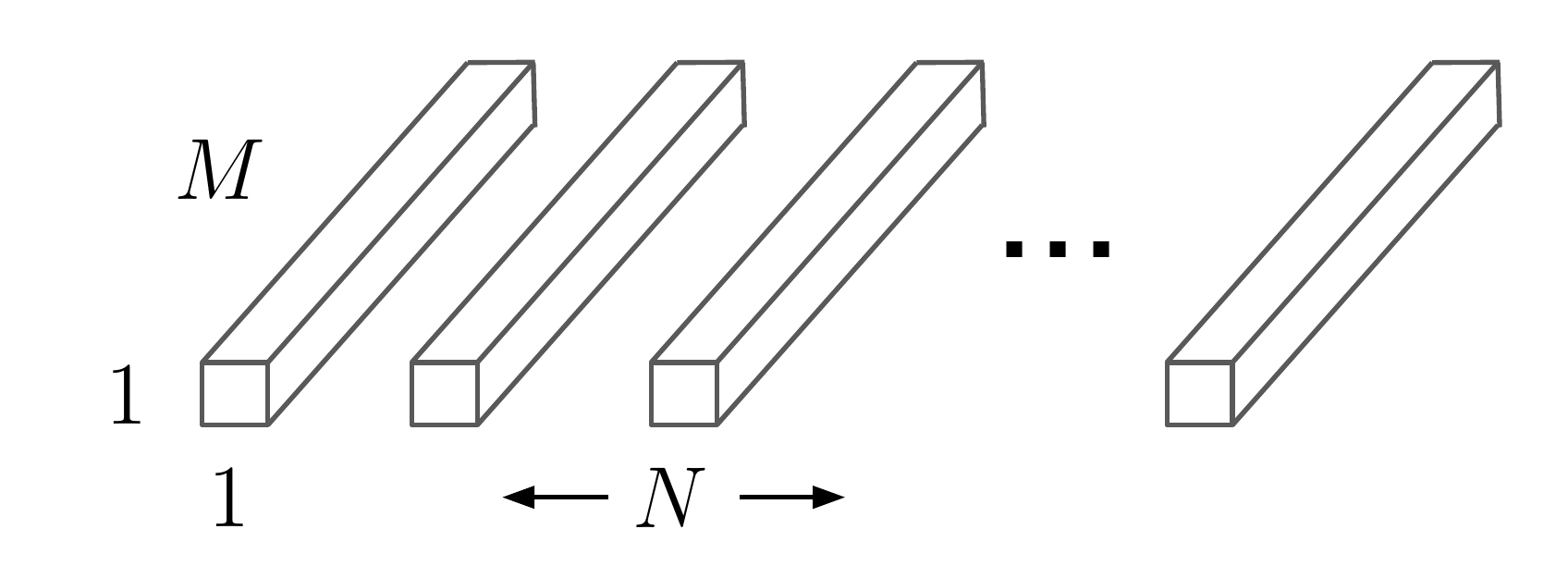}
  \label{fig:dw_conv_c}
}

\caption{The standard convolutional filters in (a) are replaced by two layers: depthwise convolution in (b) and pointwise convolution in (c) to build a depthwise separable filter.}
\label{fig:dw_conv}
\end{figure}

The MobileNet model is based on depthwise separable convolutions which is a form of factorized convolutions which factorize a standard convolution into a depthwise convolution and a $1 \times 1$ convolution called a pointwise convolution. For MobileNets the depthwise convolution applies a single filter to each input channel. The pointwise convolution then applies a $1 \times 1$ convolution to combine the outputs the depthwise convolution. A standard convolution both filters and combines inputs into a new set of outputs in one step. The depthwise separable convolution splits this into two layers, a separate layer for filtering and a separate layer for combining. This factorization has the effect of drastically reducing computation and model size. Figure \ref{fig:dw_conv} shows how a standard convolution \ref{fig:dw_conv_a} is factorized into a depthwise convolution \ref{fig:dw_conv_b} and a $1 \times 1$ pointwise convolution \ref{fig:dw_conv_c}.

A standard convolutional layer takes as input a $D_F \times D_F \times M$ feature map $\mathbf{F}$ and produces a $D_F \times D_F \times N$ feature map $\mathbf{G}$ where $D_F$ is the spatial width and height of a square input feature map\footnote{We assume that the output feature map has the same spatial dimensions as the input and both feature maps are square. Our model shrinking results generalize to feature maps with arbitrary sizes and aspect ratios.}, $M$ is the number of input channels (input depth), $D_G$ is the spatial width and height of a square output feature map and $N$ is the number of output channel (output depth).

The standard convolutional layer is parameterized by convolution kernel $\mathbf{K}$ of size $D_K \times D_K \times M \times N$ where $D_K$ is the spatial dimension of the kernel assumed to be square and $M$ is number of input channels and $N$ is the number of output channels as defined previously. 

The output feature map for standard convolution assuming stride one and padding is computed as:

\begin{equation}
\mathbf{G}_{k,l,n} = \sum_{i,j,m} \mathbf{K}_{i,j,m,n} \cdot \mathbf{F}_{k+i-1,l+j-1,m}
\end{equation}

Standard convolutions have the computational cost of:

\begin{equation}
D_K \cdot D_K \cdot M \cdot N \cdot D_F \cdot D_F
\end{equation}
where the computational cost depends multiplicatively on the number of input channels $M$, the number of output channels $N$ the kernel size $D_k \times D_k$ and the feature map size $D_F \times D_F$. MobileNet models address each of these terms and their interactions. First it uses depthwise separable convolutions to break the interaction between the number of output channels and the size of the kernel.

The standard convolution operation has the effect of filtering features based on the 
convolutional kernels and combining features in order to produce a new representation.
The filtering and combination steps can be split into two steps via the use of 
factorized convolutions called depthwise separable convolutions for substantial reduction in computational cost. 

Depthwise separable convolution are made up of two layers: depthwise convolutions and pointwise convolutions.
We use depthwise convolutions to apply a single filter per each input channel (input depth). Pointwise convolution, a
simple $1 \times 1$ convolution, is then used to create a linear combination of the output of the depthwise layer. MobileNets use both 
batchnorm and ReLU nonlinearities for both layers.

Depthwise convolution with one filter per input channel (input depth) can be written as:

\begin{equation}
\hat{\mathbf{G}}_{k,l,m} = \sum_{i,j} \hat{\mathbf{K}}_{i,j,m} \cdot \mathbf{F}_{k+i-1,l+j-1,m}
\end{equation}
where $\hat{\mathbf{K}}$ is the depthwise convolutional kernel of size $D_K \times D_K \times M$ where the $m_{th}$ filter in $\hat{\mathbf{K}}$ is applied to the $m_{th}$ channel in $\mathbf{F}$ to produce the $m_{th}$ channel of the filtered output feature map $\hat{\mathbf{G}}$.

Depthwise convolution has a computational cost of:

\begin{equation}
D_K \cdot D_K \cdot M \cdot D_F \cdot D_F
\end{equation}

Depthwise convolution is extremely efficient relative to standard convolution. However it only filters input channels, it does not combine them to create new features. So an additional layer that computes a linear combination of the output of depthwise convolution via $1 \times 1$ convolution is needed in order to generate these new features.

The combination of depthwise convolution and $1\times1$ (pointwise) convolution is called depthwise separable convolution which was originally introduced in \cite{sifre2014rigid}.

Depthwise separable convolutions cost:

\begin{equation}  
 D_K \cdot D_K \cdot M \cdot D_F \cdot D_F +  M \cdot N \cdot D_F \cdot D_F
\end{equation}
which is the sum of the depthwise and $1 \times 1$ pointwise convolutions. 

By expressing convolution as a two step process of filtering and combining we get a reduction in computation of:

\begin{eqnarray*}
&&\frac{D_K \cdot D_K \cdot M \cdot D_F \cdot D_F +  M \cdot N \cdot D_F \cdot D_F}{D_K \cdot D_K \cdot M \cdot N \cdot D_F \cdot D_F} \\
&=&\frac{1}{N} + \frac{1}{D_K^2}
\end{eqnarray*}

MobileNet uses $3 \times 3$ depthwise separable convolutions which uses between 8 to 9 times less computation than standard convolutions at only a small reduction in accuracy as seen in Section \ref{sec:exp}.

Additional factorization in spatial dimension such as in \cite{jin2014flattened,szegedy2015rethinking} does not save much additional computation as very little computation is spent in depthwise convolutions.

\subsection{Network Structure and Training}

The MobileNet structure is built on depthwise separable convolutions as mentioned in the previous section except for the first layer which is a full convolution. By defining the network in such simple terms we are able to easily explore network topologies to find a good network. The MobileNet architecture is defined in Table \ref{table:mobilenet}. All layers are followed by a batchnorm \cite{ioffe2015batch} and ReLU nonlinearity with the exception of the final fully connected layer which has no nonlinearity and feeds into a softmax layer for classification. Figure \ref{fig:conv_layers} contrasts a layer with regular convolutions, batchnorm and ReLU nonlinearity to the factorized layer with depthwise convolution, $1 \times 1$ pointwise convolution as well as batchnorm and ReLU after each convolutional layer. Down sampling is handled with strided convolution in the depthwise convolutions as well as in the first layer. A final average pooling reduces the spatial resolution to 1 before the fully connected layer. Counting depthwise and pointwise convolutions as separate layers, MobileNet has 28 layers.

It is not enough to simply define networks in terms of a small number of Mult-Adds. It is also important to make sure these operations can be efficiently implementable. For instance unstructured sparse matrix operations are not typically faster than dense matrix operations until a very high level of sparsity. Our model structure puts nearly all of the computation into dense $1 \times 1$ convolutions. This can be implemented with highly optimized general matrix multiply (GEMM) functions. Often convolutions are implemented by a GEMM but require an initial reordering in memory called im2col in order to map it to a GEMM. For instance, this approach is used in the popular Caffe package \cite{jia2014caffe}. $1 \times 1$ convolutions do not require this reordering in memory and can be implemented directly with GEMM which is one of the most optimized numerical linear algebra algorithms. MobileNet spends $95\%$ of it's computation time in $1 \times 1$ convolutions which also has $75\%$ of the parameters as can be seen in Table \ref{table:percent}. Nearly all of the additional parameters are in the fully connected layer.

MobileNet models were trained in TensorFlow \cite{abadi2015tensorflow} using RMSprop \cite{tieleman2012lecture} with asynchronous gradient descent similar to Inception V3 \cite{szegedy2015rethinking}. However, contrary to training large models we use less regularization and data augmentation techniques because small models have less trouble with overfitting. When training MobileNets we do not use side heads or label smoothing and additionally reduce the amount image of distortions by limiting the size of small crops that are used in large Inception training \cite{szegedy2015rethinking}. Additionally, we found that it was important to put very little or no weight decay (l2 regularization) on the depthwise filters since their are so few parameters in them. For the ImageNet benchmarks in the next section all models were trained with same training parameters regardless of the size of the model.  

\begin{table}[t]
  \caption{MobileNet Body Architecture} 
\centering 
\resizebox{\columnwidth}{!}{%
\begin{tabular}{l | l | l} 
\hline\hline 
Type / Stride & Filter Shape & Input Size \\ [0.5ex] 
\hline 
Conv / s2 & $3\times3\times3\times32$ & $224\times224\times3$\\
\hline
Conv dw / s1& $3\times3\times32$ dw & $112\times112\times32$\\
\hline
Conv / s1& $1\times1\times32\times64$ & $112\times112\times32$\\
\hline
Conv dw / s2& $3\times3\times64$ dw & $112\times112\times64$\\
\hline
Conv / s1& $1\times1\times64\times128$ & $56\times56\times64$\\
\hline
Conv dw / s1& $3\times3\times128$ dw & $56\times56\times128$\\
\hline
Conv / s1& $1\times1\times128\times128$ & $56\times56\times128$\\
\hline
Conv dw / s2& $3\times3\times128$ dw & $56\times56\times128$\\
\hline
Conv / s1& $1\times1\times128\times256$ & $28\times28\times128$\\
\hline
Conv dw / s1& $3\times3\times256$ dw & $28\times28\times256$\\
\hline
Conv / s1& $1\times1\times256\times256$ & $28\times28\times256$\\
\hline
Conv dw / s2& $3\times3\times256$ dw & $28\times28\times256$\\
\hline
Conv / s1& $1\times1\times256\times512$ & $14\times14\times256$\\
\hline
\multirow{2}{*}{$5\times$} Conv dw / s1& $3\times3\times512$ dw & $14\times14\times512$\\
\hspace{.53cm}Conv / s1& $1\times1\times512\times512$ & $14\times14\times512$\\
\hline
Conv dw / s2& $3\times3\times512$ dw & $14\times14\times512$\\
\hline
Conv / s1& $1\times1\times512\times1024$ & $7\times7\times512$\\
\hline
Conv dw / s2& $3\times3\times1024$ dw & $7\times7\times1024$\\
\hline
Conv / s1& $1\times1\times1024\times1024$ & $7\times7\times1024$\\
\hline
Avg Pool / s1& Pool $7\times7$ & $7\times7\times1024$\\
\hline
FC / s1 & $1024 \times 1000$ & $1\times1\times1024$\\
\hline
Softmax / s1 & Classifier & $1\times1\times1000$\\
\hline 
\end{tabular}
\label{table:mobilenet} 
}
\end{table}

\begin{figure}
  \centering
  \includegraphics[width=0.8\linewidth]{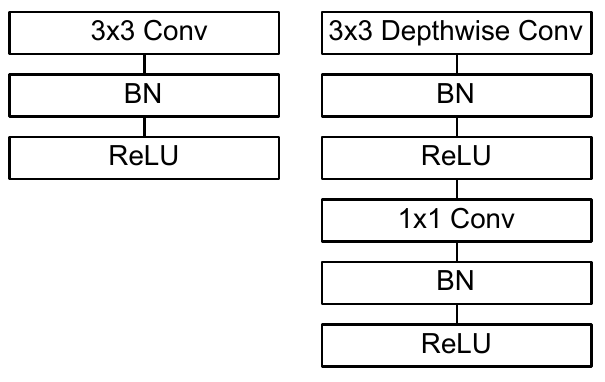}
  \caption{Left: Standard convolutional layer with batchnorm and ReLU. Right: Depthwise Separable convolutions with Depthwise and Pointwise layers followed by batchnorm and ReLU.}
  \label{fig:conv_layers}
\end{figure}

\begin{table}[t]
  \caption{Resource Per Layer Type} 
\centering 
\begin{tabular}{l | l | l} 
\hline\hline 
 Type & Mult-Adds  & Parameters \\ [0.5ex] 
\hline 
Conv $1 \times 1$ & 94.86\% & 74.59\%   \\ 
\hline
Conv DW $3 \times 3$ & 3.06\% & 1.06\% \\
\hline
Conv $3 \times 3$ & 1.19\% & 0.02\% \\
\hline
Fully Connected & 0.18\% & 24.33\% \\
[1ex] 
\hline 
\end{tabular}
\label{table:percent} 
\end{table}

\subsection{Width Multiplier: Thinner Models}

Although the base MobileNet architecture is already small and low latency, many times a specific use case or application may require the model to be smaller and faster. In order to construct these smaller and less computationally expensive models we introduce a very simple parameter $\alpha$ called width multiplier. The role of the width multiplier $\alpha$ is to thin a network uniformly at each layer. For a given layer and width multiplier $\alpha$, the number of input channels $M$ becomes $\alpha M$ and the number of output channels $N$ becomes $\alpha N$.

The computational cost of a depthwise separable convolution with width multiplier $\alpha$ is:  
\begin{equation}
D_K \cdot D_K \cdot \alpha M \cdot D_F \cdot D_F +  \alpha M \cdot \alpha N \cdot D_F \cdot D_F
\end{equation}
where $\alpha \in (0,1]$ with typical settings of 1, 0.75, 0.5 and 0.25. $\alpha=1$ is the baseline MobileNet and $\alpha<1$
are reduced MobileNets. Width multiplier has the effect of reducing computational cost and the number of parameters quadratically by roughly $\alpha^2$. Width multiplier can be applied to any model structure to define a new smaller model with a reasonable accuracy, latency and size trade off. It is used to define a new reduced structure that needs to be trained from scratch.

\subsection{Resolution Multiplier: Reduced Representation}

The second hyper-parameter to reduce the computational cost of a neural network is a resolution multiplier $\rho$. We apply this to the input image and the internal representation of every layer is subsequently reduced by the same multiplier. In practice we implicitly set $\rho$ by setting the input resolution. 

We can now express the computational cost for the core layers of our network as depthwise separable convolutions with width multiplier $\alpha$ and resolution multiplier $\rho$:
\begin{equation}
D_K \cdot D_K \cdot \alpha M \cdot \rho D_F \cdot \rho D_F +  \alpha M \cdot \alpha N \cdot \rho D_F \cdot \rho D_F
\end{equation}
where $\rho \in (0,1]$ which is typically set implicitly so that the input resolution of the network is 224, 192, 160 or 128. $\rho=1$ is the baseline MobileNet and $\rho<1$
are reduced computation MobileNets. Resolution multiplier has the effect of reducing computational cost by $\rho^2$.

As an example we can look at a typical layer in MobileNet and see how depthwise separable convolutions, width multiplier and resolution multiplier reduce the cost and parameters. Table \ref{table:layer_resource} shows the computation and number of parameters for a layer as architecture shrinking methods are sequentially applied to the layer. The first row shows the Mult-Adds and parameters for a full convolutional layer with an input feature map of size $14 \times 14 \times 512$ with a kernel $K$ of size $3 \times 3 \times 512 \times 512$. We will look in detail in the next section at the trade offs between resources and accuracy.

\begin{table}[t]
  \caption{Resource usage for modifications to standard convolution. Note that each row is a cumulative effect adding on top of the previous row. This example is for an internal MobileNet layer with $D_K=3$, $M=512$, $N=512$, $D_F=14$.} 
\centering 
\resizebox{\columnwidth}{!}{%
\begin{tabular}{c c c c} 
\hline\hline 
Layer/Modification &  Million & Million \\ [0.5ex] 
 & Mult-Adds & Parameters \\
\hline 
Convolution & 462 & 2.36 \\ 
Depthwise Separable Conv & 52.3 & 0.27 \\
$\alpha=0.75$ & 29.6 & 0.15 \\
$\rho=0.714$ & 15.1 & 0.15 \\
 [1ex] 
\hline 
\end{tabular}
\label{table:layer_resource} 
}
\end{table}

\section{Experiments} \label{sec:exp}

In this section we first investigate the effects of depthwise convolutions as well as the choice of shrinking by reducing the width of the network rather than the number of layers. We then show the trade offs of reducing the network based on the two hyper-parameters: width multiplier and resolution multiplier and compare results to a number of popular models. We then investigate MobileNets applied to a number of different applications.

\subsection{Model Choices}
First we show results for MobileNet with depthwise separable convolutions compared to a model built with full convolutions. In Table \ref{table:dm_full} we see that using depthwise separable convolutions compared to full convolutions only reduces accuracy by $1\%$ on ImageNet was saving tremendously on mult-adds and parameters.

\begin{table}[t]
  \caption{Depthwise Separable vs Full Convolution MobileNet} 
\centering 
\resizebox{\columnwidth}{!}{%
\begin{tabular}{c c c c} 
\hline\hline 
Model & ImageNet & Million & Million \\ [0.5ex] 
 &  Accuracy & Mult-Adds & Parameters \\
\hline 
Conv MobileNet & 71.7\% & 4866 & 29.3\\
MobileNet & 70.6\% & 569 & 4.2 \\ 
 [1ex] 
\hline 
\end{tabular}
\label{table:dm_full} 
}
\end{table}

We next show results comparing thinner models with width multiplier to shallower models using less layers. To make MobileNet shallower, the $5$ layers of separable filters with feature size $14 \times 14 \times 512$ in Table \ref{table:mobilenet} are removed. Table \ref{table:dm_shallow} shows that at similar computation and number of parameters, that making MobileNets thinner is $3\%$ better than making them shallower.

\begin{table}[t]
  \caption{Narrow vs Shallow MobileNet} 
\centering 
\resizebox{\columnwidth}{!}{%
\begin{tabular}{c c c c} 
\hline\hline 
Model & ImageNet & Million & Million \\ [0.5ex] 
 &  Accuracy & Mult-Adds & Parameters \\
\hline 
0.75 MobileNet & 68.4\% & 325 & 2.6 \\ 
Shallow MobileNet & 65.3\% & 307 & 2.9\\
 [1ex] 
\hline 
\end{tabular}
\label{table:dm_shallow} 
}
\end{table}

\subsection{Model Shrinking Hyperparameters}

Table \ref{table:wm} shows the accuracy, computation and size trade offs of shrinking the MobileNet architecture with the width multiplier $\alpha$.
Accuracy drops off smoothly until the architecture is made too small at $\alpha=0.25$.

\begin{table}[t]
  \caption{MobileNet Width Multiplier} 
\centering 
\resizebox{\columnwidth}{!}{%
\begin{tabular}{c c c c} 
\hline\hline 
Width Multiplier & ImageNet & Million & Million \\ [0.5ex] 
 &  Accuracy & Mult-Adds & Parameters \\
\hline 
1.0 MobileNet-224 & 70.6\% & 569 & 4.2 \\ 
0.75 MobileNet-224 & 68.4\% & 325 & 2.6 \\
0.5 MobileNet-224 & 63.7\% & 149 & 1.3 \\
0.25 MobileNet-224 & 50.6\% & 41 & 0.5 \\ [1ex] 
\hline 
\end{tabular}
\label{table:wm} 
}
\end{table}

Table \ref{table:rm} shows the accuracy, computation and size trade offs for different resolution multipliers by
training MobileNets with reduced input resolutions. Accuracy drops off smoothly across resolution.

\begin{table}[t]
  \caption{MobileNet Resolution} 
\centering 
\resizebox{\columnwidth}{!}{%
\begin{tabular}{c c c c} 
\hline\hline 
Resolution & ImageNet & Million & Million \\ [0.5ex] 
 &  Accuracy & Mult-Adds & Parameters \\
\hline 
1.0 MobileNet-224 & 70.6\% & 569  & 4.2 \\ 
1.0 MobileNet-192 & 69.1\% & 418 & 4.2 \\
1.0 MobileNet-160 & 67.2\% & 290 & 4.2 \\
1.0 MobileNet-128 & 64.4\% & 186 & 4.2 \\ [1ex] 
\hline 
\end{tabular}
\label{table:rm} 
}
\end{table}

\begin{figure}
  \includegraphics[width=\linewidth]{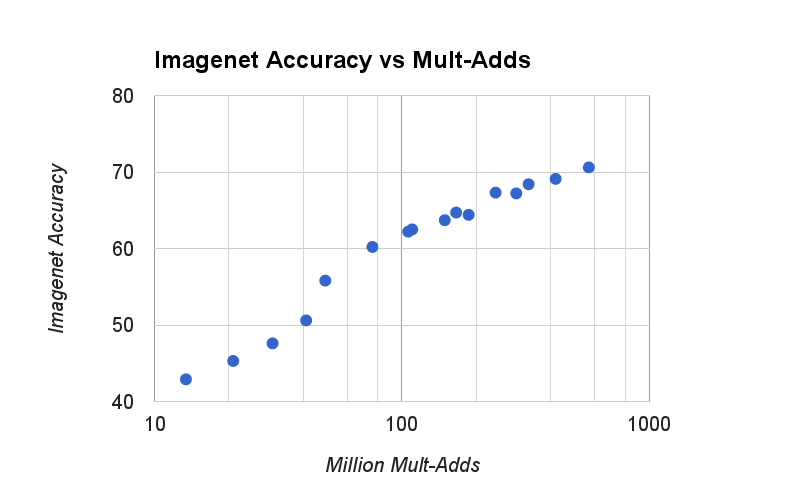}
  \caption{This figure shows the trade off between computation (Mult-Adds) and accuracy on the ImageNet benchmark. Note the log linear dependence between accuracy and computation.}
  \label{fig:mult-add}
\end{figure}

\begin{figure}
  \includegraphics[width=\linewidth]{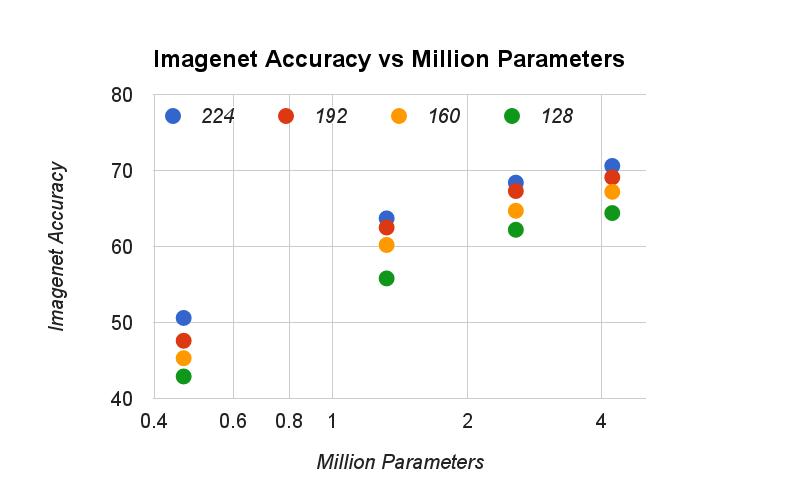}
  \caption{This figure shows the trade off between the number of parameters and accuracy on the ImageNet benchmark. The colors encode input resolutions. The number of parameters do not vary based on the input resolution. }
  \label{fig:parameters}
\end{figure}

Figure \ref{fig:mult-add} shows the trade off between ImageNet Accuracy and computation for the 16 models made from the cross product of width multiplier $\alpha \in \{1,0.75,0.5,0.25\}$ and resolutions $\{224, 192, 160, 128\}$. Results are log linear with a jump when models get very small at $\alpha=0.25$.

Figure \ref{fig:parameters} shows the trade off between ImageNet Accuracy and number of parameters for the 16 models made from the cross product of width multiplier $\alpha \in \{1,0.75,0.5,0.25\}$ and resolutions $\{224, 192, 160, 128\}$. 

Table \ref{table:mncompare} compares full MobileNet to the original GoogleNet \cite{szegedy2015going} and VGG16 \cite{simonyan2014very}. MobileNet is nearly as accurate as VGG16 while being 32 times smaller and 27 times less compute intensive. It is more accurate than GoogleNet while being smaller and more than 2.5 times less computation.

Table \ref{table:mncompare2} compares a reduced MobileNet with width multiplier $\alpha=0.5$ and reduced resolution $160\times160$. Reduced MobileNet is $4\%$ better than AlexNet \cite{krizhevsky2012imagenet} while being $45\times$ smaller and $9.4\times$ less compute than AlexNet. It is also $4\%$ better than Squeezenet \cite{iandola2016squeezenet} at about the same size and $22 \times$ less computation.

\begin{table}[t]
  \caption{MobileNet Comparison to Popular Models} 
\centering 
\resizebox{\columnwidth}{!}{%
\begin{tabular}{c c c c} 
\hline\hline 
Model & ImageNet & Million & Million \\ [0.5ex] 
 &  Accuracy & Mult-Adds & Parameters \\
\hline 
1.0 MobileNet-224 & 70.6\% & 569 & 4.2 \\ 
GoogleNet & 69.8\% & 1550 & 6.8 \\
VGG 16 & 71.5\% & 15300 & 138 \\ [1ex] 
\hline 
\end{tabular}
\label{table:mncompare} 
}
\end{table}

\begin{table}[t]
  \caption{Smaller MobileNet Comparison to Popular Models} 
\centering 
\resizebox{\columnwidth}{!}{%
\begin{tabular}{c c c c} 
\hline\hline 
Model & ImageNet & Million & Million \\ [0.5ex] 
 &  Accuracy & Mult-Adds & Parameters \\
\hline 
0.50 MobileNet-160 & 60.2\% & 76 & 1.32 \\
Squeezenet & 57.5\% & 1700 & 1.25 \\
AlexNet & 57.2\% & 720 & 60 \\ [1ex] 
\hline 
\end{tabular}
\label{table:mncompare2} 
}
\end{table}

\subsection{Fine Grained Recognition}
We train MobileNet for fine grained recognition on the Stanford Dogs dataset \cite{KhoslaYaoJayadevaprakashFeiFei_FGVC2011}.
We extend the approach of \cite{krause2015unreasonable} and collect an even larger but noisy training set than \cite{krause2015unreasonable} from the web.
We use the noisy web data to pretrain a fine grained dog recognition model and then fine tune the model on the Stanford Dogs training set.
Results on Stanford Dogs test set are in Table \ref{table:dogs}. MobileNet can almost achieve the state of the art results from
\cite{krause2015unreasonable} at greatly reduced computation and size.

\begin{table}[t]
  \caption{MobileNet for Stanford Dogs} 
\centering 
\resizebox{\columnwidth}{!}{%
\begin{tabular}{c c c c} 
\hline\hline 
Model & Top-1 & Million & Million \\ [0.5ex] 
 & Accuracy & Mult-Adds & Parameters \\
\hline 
Inception V3 \cite{krause2015unreasonable} & 84\% & 5000 & 23.2 \\ 
1.0 MobileNet-224 & 83.3\% & 569 & 3.3 \\
0.75 MobileNet-224 & 81.9\% & 325 & 1.9 \\
1.0 MobileNet-192 & 81.9\% &  418 & 3.3 \\
0.75 MobileNet-192 & 80.5\% & 239 & 1.9 \\[1ex] 
\hline 
\end{tabular}
\label{table:dogs} 
}
\end{table}

\subsection{Large Scale Geolocalizaton}
\begin{table}[t]
\setlength\tabcolsep{3pt}
  \caption{Performance of PlaNet using the MobileNet architecture. Percentages are the fraction of the Im2GPS test dataset that were localized within a certain distance from the ground truth. The numbers for the original PlaNet model are based on an updated version that has an improved architecture and training dataset.} 
\centering 
\resizebox{\columnwidth}{!}{%
\begin{tabular}{c c c c} 
\hline\hline 
Scale & Im2GPS \cite{hays2014large} & PlaNet \cite{weyand2016planet} & PlaNet \\
      &  &   & MobileNet \\
\hline 
\small{Continent (2500 km)} & 51.9\% & 77.6\% & 79.3\% \\ 
\small{Country (750 km)}    & 35.4\% & 64.0\% & 60.3\% \\
\small{Region (200 km)}     & 32.1\% & 51.1\% & 45.2\% \\
\small{City (25 km)}        & 21.9\% & 31.7\% & 31.7\% \\
\small{Street (1 km)}       &  2.5\% & 11.0\% & 11.4\% \\ [1ex] 
\hline 
\end{tabular}
\label{table:planet} 
}
\end{table}
PlaNet \cite{weyand2016planet} casts the task of determining where on earth a photo was taken as a classification problem. The approach divides the earth into a grid of geographic cells that serve as the target classes and trains a convolutional neural network on millions of geo-tagged photos. PlaNet has been shown to successfully localize a large variety of photos and to outperform Im2GPS \cite{hays2008im2gps,hays2014large} that addresses the same task.

We re-train PlaNet using the MobileNet architecture on the same data. While the full PlaNet model based on the Inception V3 architecture \cite{szegedy2015rethinking} has 52 million parameters and 5.74 billion mult-adds. The MobileNet model has only 13 million parameters with the usual 3 million for the body and 10 million for the final layer and 0.58 Million mult-adds. As shown in Tab.~\ref{table:planet}, the MobileNet version delivers only slightly decreased performance compared to PlaNet despite being much more compact. Moreover, it still outperforms Im2GPS by a large margin.

\subsection{Face Attributes}
Another use-case for MobileNet is compressing large systems with unknown or esoteric training procedures. In a face attribute classification task, we demonstrate a synergistic relationship between MobileNet and distillation~\cite{hinton2015distilling}, a knowledge transfer technique for deep networks. We seek to reduce a large face attribute classifier with $75$ million parameters and $1600$ million Mult-Adds. The classifier is trained on a multi-attribute dataset similar to YFCC100M~\cite{thomee2016yfcc100m}.

We distill a face attribute classifier using the MobileNet architecture. Distillation~\cite{hinton2015distilling} works by training the classifier to emulate the outputs of a larger model\footnote{The emulation quality is measured by averaging the per-attribute cross-entropy over all attributes.} instead of the ground-truth labels, hence enabling training from large (and potentially infinite) unlabeled datasets. Marrying the scalability of distillation training and the parsimonious parameterization of MobileNet, the end system not only requires no regularization (e.g. weight-decay and early-stopping), but also demonstrates enhanced performances. It is evident from Tab.~\ref{table:faceattr} that the MobileNet-based classifier is resilient to aggressive model shrinking: it achieves a similar mean average precision across attributes (mean AP) as the in-house while consuming only $1\%$ the Multi-Adds. 

\begin{table}[t]
\setlength\tabcolsep{3pt}
  \caption{Face attribute classification using the MobileNet architecture. Each row corresponds to a different hyper-parameter setting (width multiplier $\alpha$ and image resolution).} 
\centering 
\resizebox{\columnwidth}{!}{%
\begin{tabular}{c c c c} 
\hline\hline 
Width Multiplier / & Mean  & Million & Million \\
   Resolution   & AP  & Mult-Adds  & Parameters \\
\hline 
1.0 MobileNet-224 & 88.7\% & 568 & 3.2 \\ 
0.5 MobileNet-224    & 88.1\% & 149 & 0.8 \\
0.25 MobileNet-224   & 87.2\% & 45 & 0.2 \\
1.0 MobileNet-128  & 88.1\% & 185 & 3.2 \\
0.5 MobileNet-128 &  87.7\% & 48 & 0.8 \\
0.25 MobileNet-128 &  86.4\% & 15 & 0.2 \\
\hline 
Baseline & 86.9\% & 1600 & 7.5 \\
\end{tabular}
\label{table:faceattr} 
}
\end{table}

\subsection{Object Detection}

MobileNet can also be deployed as an effective base network in modern object detection systems.
We report results for MobileNet trained for object detection on COCO data based on the recent work that won the 2016 COCO challenge \cite{cocodetection2016}.
In table \ref{table:objectdetection}, MobileNet is compared to VGG and Inception V2 \cite{ioffe2015batch} under both Faster-RCNN \cite{ren2015faster} and SSD \cite{liu2015ssd} framework.
In our experiments, SSD is evaluated with 300 input resolution (SSD 300) and Faster-RCNN is compared with both 300 and 600 input resolution (Faster-RCNN 300, Faster-RCNN 600).
The Faster-RCNN model evaluates 300 RPN proposal boxes per image. The models are trained on COCO train+val excluding 8k minival images and evaluated on minival.
For both frameworks, MobileNet achieves comparable results to other networks with only a fraction of computational complexity and model size.

\begin{table}[t]
\setlength\tabcolsep{3pt}
  \caption{COCO object detection results comparison using different frameworks and network architectures. mAP is reported with COCO primary challenge metric (AP at IoU=0.50:0.05:0.95)} 
\centering 
\resizebox{\columnwidth}{!}{%
\begin{tabular}{c c c c c} 
\hline\hline 
Framework  & Model & mAP & Billion    & Million \\
Resolution &       &     & Mult-Adds  & Parameters \\
\hline 
        & deeplab-VGG  & 21.1\% & 34.9 & 33.1 \\ 
SSD 300 & Inception V2 & 22.0\% & 3.8  & 13.7 \\ 
        & MobileNet    & 19.3\% & 1.2  & 6.8  \\
\hline 
Faster-RCNN & VGG          & 22.9\% & 64.3  & 138.5 \\
300         & Inception V2 & 15.4\% & 118.2 & 13.3 \\
            & MobileNet    & 16.4\% & 25.2  & 6.1  \\
\hline 
Faster-RCNN & VGG          & 25.7\% & 149.6 & 138.5 \\
600         & Inception V2 & 21.9\% & 129.6 & 13.3 \\
            & Mobilenet    & 19.8\% & 30.5  & 6.1  \\
\hline 
\end{tabular}
}
\label{table:objectdetection} 
\end{table}

\begin{figure}
  \includegraphics[width=\linewidth]{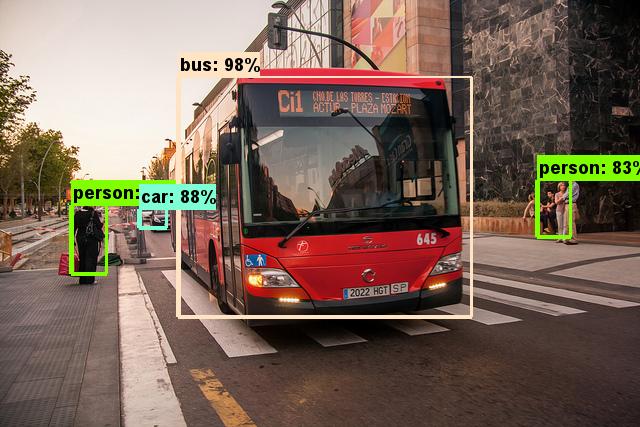}
  \caption{Example objection detection results using MobileNet SSD.}
  \label{fig:detection}
\end{figure}

\subsection{Face Embeddings}
The FaceNet model is a state of the art face recognition model \cite{schroff2015facenet}. It builds face embeddings based on the triplet loss. To build a mobile FaceNet model we use distillation
to train by minimizing the squared differences of the output of FaceNet and MobileNet on the training data. Results for very small MobileNet models can be found in table \ref{table:facenet}.

\begin{table}[t]
  \caption{MobileNet Distilled from FaceNet} 
\centering 
\resizebox{\columnwidth}{!}{%
\begin{tabular}{c c c c} 
\hline\hline 
Model & 1e-4 & Million & Million \\ [0.5ex] 
 &  Accuracy & Mult-Adds & Parameters \\
\hline 
FaceNet \cite{schroff2015facenet} & 83\% & 1600 & 7.5 \\ 
1.0 MobileNet-160 & 79.4\% & 286 & 4.9 \\
1.0 MobileNet-128 & 78.3\% & 185 & 5.5 \\
0.75 MobileNet-128 & 75.2\% & 166 & 3.4 \\
0.75 MobileNet-128 & 72.5\% & 108 &  3.8 \\
[1ex] 
\hline 
\end{tabular}
\label{table:facenet} 
}
\end{table}

\section{Conclusion} \label{sec:conclusion}

We proposed a new model architecture called MobileNets based on depthwise separable convolutions. We investigated some of the important design decisions leading to an efficient model. We then demonstrated how to build smaller and faster MobileNets using width multiplier and resolution multiplier by trading off a reasonable amount of accuracy to reduce size and latency. We then compared different MobileNets to popular models demonstrating superior size, speed and accuracy characteristics. We concluded by demonstrating MobileNet's effectiveness when applied to a wide variety of tasks. As a next step to help adoption and exploration of MobileNets, we plan on releasing models in Tensor Flow.

{\small
\bibliographystyle{ieee}
\bibliography{mobilenet_bib}
}

\end{document}